\documentclass[letterpaper, 10 pt, conference]{ieeeconf} 

\IEEEoverridecommandlockouts

\usepackage{graphics} 
\usepackage{graphicx,import} 
\usepackage{epsfig} 
\usepackage{mathptmx} 
\usepackage{amsmath} 
\usepackage{amssymb}  
\usepackage{mathtools}
\DeclarePairedDelimiterX\set[1]\lbrace\rbrace{#1}
\usepackage{hyperref}
\hypersetup{
    colorlinks=true,
    linkcolor=blue,
    filecolor=magenta,      
    urlcolor=blue,
}
\title{\LARGE \bf
Training in Task Space to Speed Up and Guide Reinforcement Learning
}

\author{Guillaume Bellegarda and Katie Byl%
\thanks{This work was funded in part by NSF NRI award 1526424.}
\thanks{Guillaume Bellegarda and Katie Byl are with the Robotics Laboratory, Department of Electrical and Computer Engineering, University of California at Santa Barbara (UCSB).
        {\tt\small gbellegarda@ucsb.edu, katiebyl@ucsb.edu}}%
}

\begin{document}
\maketitle
\thispagestyle{empty}
\pagestyle{empty}

\begin{abstract}

Recent breakthroughs in the reinforcement learning (RL) community have made significant advances towards learning and deploying policies on real world robotic systems.
However, even with the current state-of-the-art algorithms and computational resources, these algorithms are still plagued with high sample complexity, and thus long training times, especially for high degree of freedom (DOF) systems.
There are also concerns arising from lack of perceived stability or robustness guarantees from emerging policies.
This paper aims at mitigating these drawbacks by:
(1) modeling a complex, high DOF system with a representative simple one,
(2) making explicit use of forward and inverse kinematics without forcing the RL algorithm to ``learn'' them on its own, and
(3) learning locomotion policies in Cartesian space instead of joint space.
In this paper these methods are applied to JPL's Robosimian, but can be readily used on any system with a base and end effector(s).
These locomotion policies can be produced in just a few minutes, trained on a single laptop.
We compare the robustness of the resulting learned policies to those of other control methods. An accompanying video for this paper can be found at \url{https://youtu.be/xDxxSw5ahnc}.
\end{abstract}

\section{INTRODUCTION}
\label{sec:intro}

\begin{figure}[thpb]
      \centering
      \includegraphics[width=3.4in]{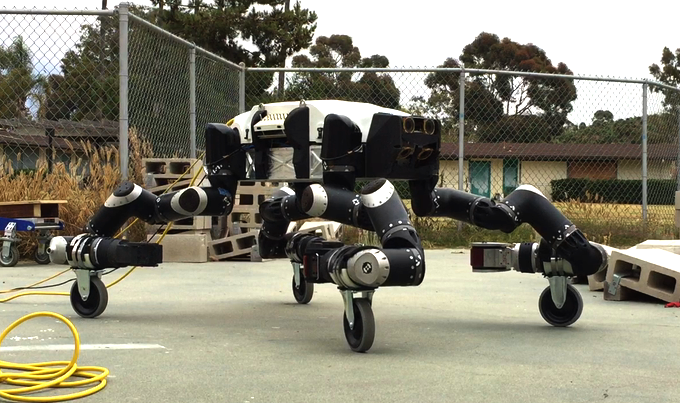} \\
      \includegraphics[width=1.7in]{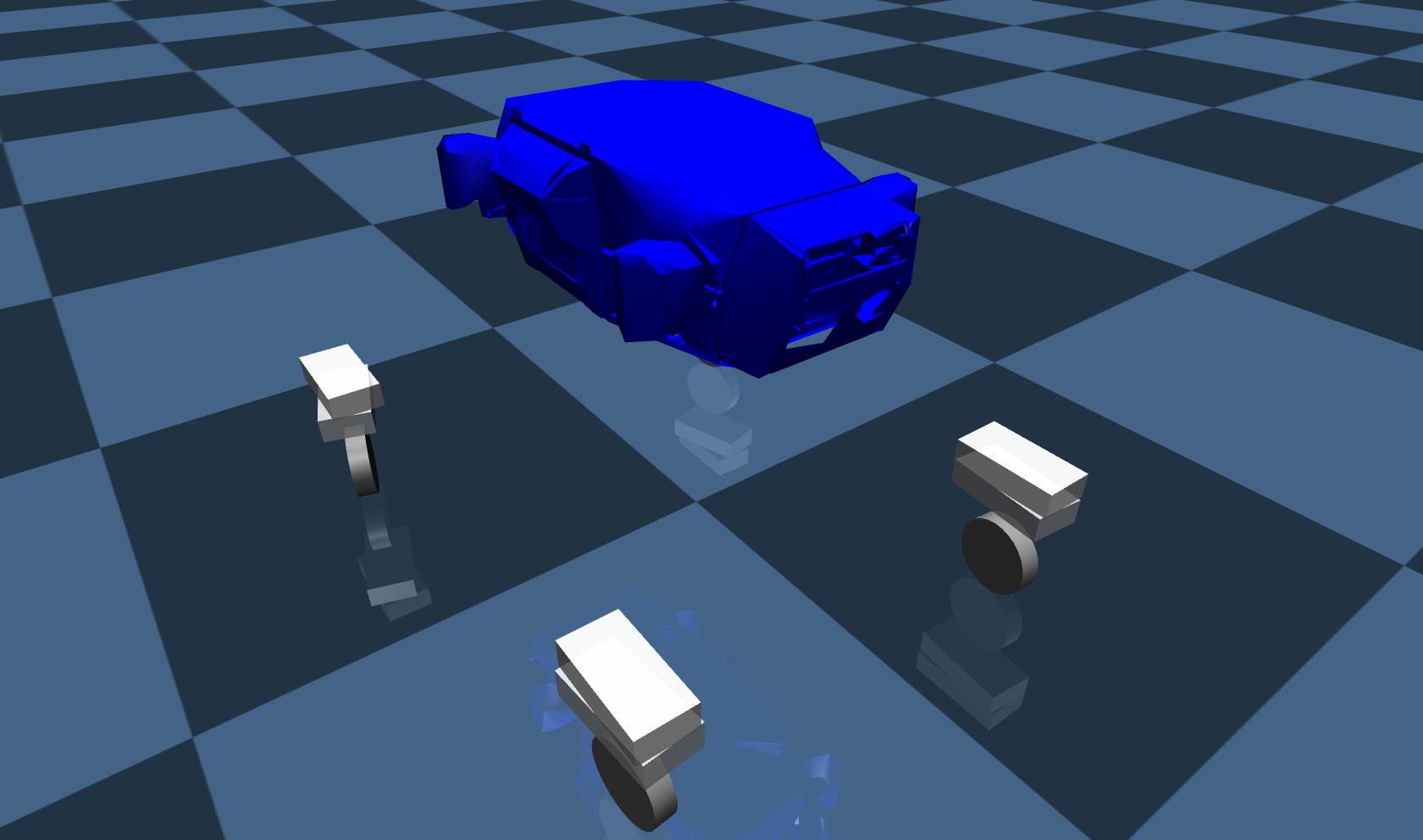}\includegraphics[width=1.7in]{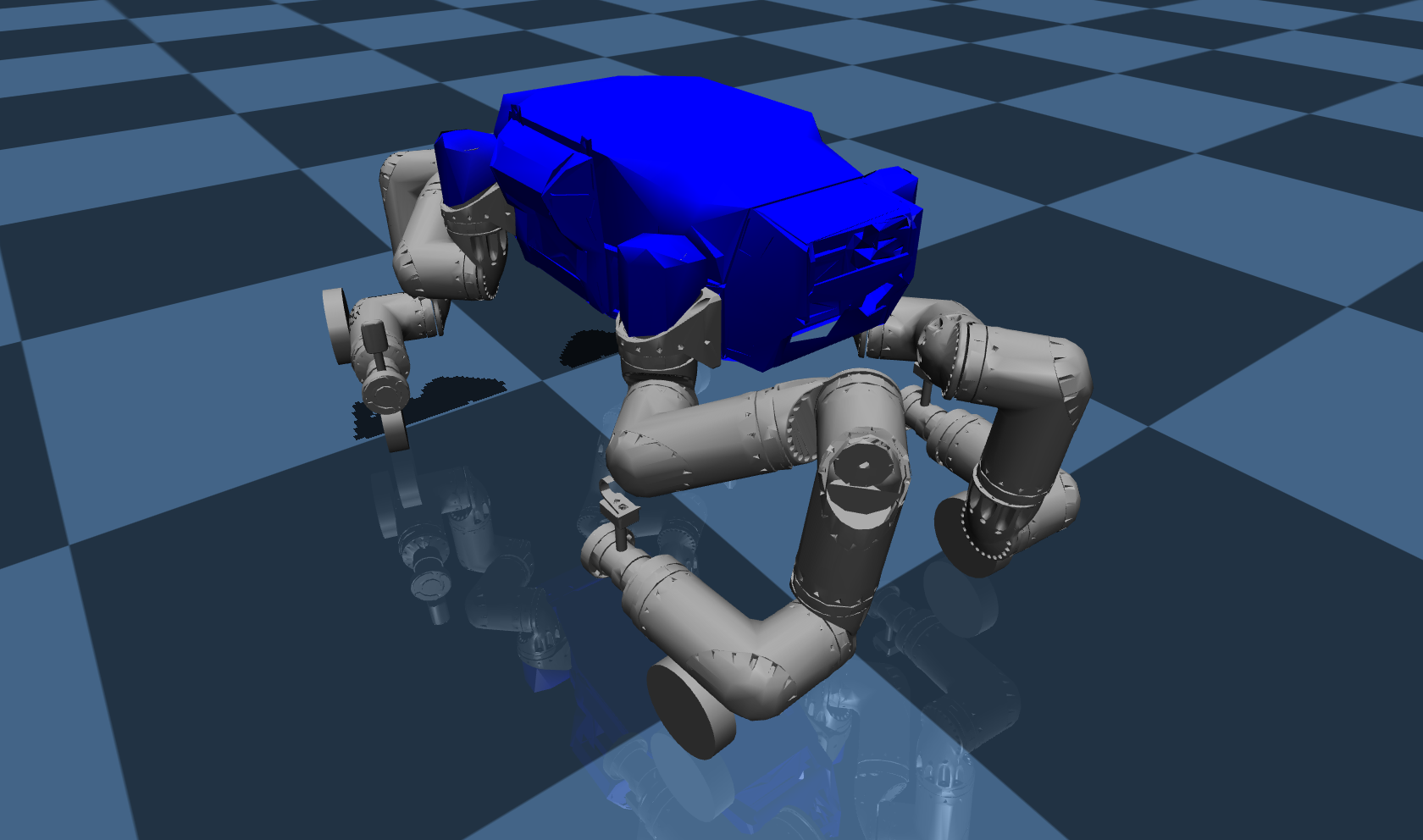}
      \caption{Versatile Locomotion with Robosimian. Top: Skating on flat ground with the real system. Bottom: Simple Cartesian space model and full Robosimian model, in MuJoCo~\cite{mujoco}.}
      \label{fig_robosimianDev}
\end{figure}

Without loss of generality to other systems with end effectors, this work aims specifically at increasing robustness and stability of skating motions designed for JPL's Robosimian quadruped~\cite{Satzinger14,Satzinger15,Byl14,Hebert15,karumanchi2017team}, which is shown in Figure~\ref{fig_robosimianDev}. Previous work in~\cite{skating2018icra} describes an overview of hand-designed skating motions on passive unactuated wheels mounted at each forearm of Robosimian's four identical limbs, comparing specifically skating with three vs. four wheels in contact with the ground. 
Results showed that on flat ground, skating on four wheels demonstrated greater robustness due to symmetry and thus decreased wheel slip.
However on terrain with bumps or other curvature, the asymmetry in configuration and contact force distribution over time from skating on three wheels had the advantage of guaranteeing continuous ground contact for all skates.
It is clear in either case that such hand-designed open-loop trajectories leave much to be desired in terms of robustness to disturbances and noise in the environment. We would like guarantees on performance, and ideally to have some confidence estimate on performance/time to complete a task in a new scenario. 

This lack of robustness, combined with impressive recent results applying reinforcement learning algorithms 
such as Proximal Policy Optimization (PPO)~\cite{googleEmergence}~\cite{ppo}, Trust Region Policy Optimization (TRPO)~\cite{trpo}, Actor Critic using Kronecker-Factored Trust Region (ACKTR)~\cite{acktr}, Deep Deterministic Policy Gradients (DDPG)~\cite{lillicrap15}, and Asynchronous Advantage Actor-Critic (A3C)~\cite{a3c}, 
to continuous control tasks in robotics,
suggests the use of (deep) reinforcement learning as a way to increase skating stability and robustness.
However, as powerful and promising as these recent results have been, the sample complexity and training time of these methods remains a major issue when seeking to deploy solutions in real time in the real world. Even for state-of-the-art algorithms and implementations, especially for high degree of freedom (DOF) complex systems such as Robosimian, a policy can take millions of iterations to train for a solution that may or may not be stable.  It must also be noted that there are no robustness, stability, or performance guarantees on the policies learned, or at least no way to readily quantify these metrics.

One prominent example is shown in the video associated with Heess et. al's \emph{Emergence of Locomotion Behaviours in Rich Environments}~\cite{googleEmergence}, where for the higher DOF system humanoid, we see the emergence of (probably) non-optimal and non-intuitive arm-flailing to ``help'' locomote the system, as a probable local optimum. The video accompanying the present paper shows an example of the emergence of similar non-intuitive
behavior when learning a locomotion policy with PPO~\cite{ppo} for Robosimian in joint space. 
We seek to avoid such local optima for locomotion policies in our system, and propose intuitively limiting the action space for reinforcement learning algorithms towards quickly generating robust and stable motions. 

Most of this recent work in applying reinforcement learning to robotic systems seeks to learn a policy that, given an observation of the current state, outputs raw motor torques to the available actuators in joint space to maximize rewards for the task at hand. 
However, for an overactuated system such as Robosimian, which has 28 actuators with high (160:1) gear ratios as well as velocity limits of 1~rad/sec at each joint~\cite{Hebert15}, applying a torque from a learned distribution at each time step is not intuitive. Rather, each motor is modeled as a position actuator, and supplied with a reference position at each time step. This naturally extends to, instead of selecting a torque at each time step, incrementing each motor's current desired position by $\Delta \in \set{-\varepsilon,0,+\varepsilon}$.

We also note that these recent learning algorithms (proudly) incorporate no prior knowledge of the system during training, and thus the agent must essentially ``learn'' forward and inverse kinematics through interacting with its environment, early termination conditions specified by a human, and hand-crafted reward shaping functions. 

Work in imitation learning, transfer learning, and warm-starting the policy network, either with existing trajectories or using more traditional controllers, has been done to try to reduce the high sample complexity of the vanilla reinforcement learning methods. We propose a much simpler idea of incorporating control techniques readily available for most systems such as forward and inverse kinematics, in the spirit that we should use the domain knowledge of the problem that we possess, rather than requiring the system to learn it on its own.

The rest of this paper is organized as follows. 
Section~\ref{sec:background} provides an overview of reinforcement learning and Proximal Policy Optimization (PPO).
Sections~\ref{sec:modeling} and \ref{sec:details} describe modeling and training environment details, respectively.
Section~\ref{sec:results} presents results for tasks such as skating with maximum velocity or to a goal location over noisy terrain, and
a brief conclusion is given in Section~\ref{sec:conclusion}.

\section{Background}
\label{sec:background}

\subsection{Reinforcement Learning}

The reinforcement learning framework,
which is described thoroughly by Sutton and Barto~\cite{sutton_rl_book} and elsewhere, typically consists of an agent interacting with an environment modeled as a Markov Decision Process (MDP). An MDP is given by a 4-tuple $(S,A,T,R)$, where $\emph{S}$ is the set of states, $\emph{A}$ is the set of actions available to the agent, $T: S \times A \times S \rightarrow \mathbb{R}$ is the transition function, where $T(s,a,s')$ gives the probability of being in state $s$, taking action $a$, and ending up in state $s'$, and  $R: S \times A \times S \rightarrow \mathbb{R}$ is the reward function, where $R(s,a,s')$ gives the expected reward for being in state $s$, taking action $a$, and ending up in state $s'$.
The goal of an agent is thus  to interact with the environment by selecting actions that will maximize future rewards.

In this paper, the states consist of a subset of the robot's (Robosimian's) positions and velocities, the actions are motor positions or Cartesian coordinate end effector offsets, the transition function is modeled by a physics engine (MuJoCo~\cite{mujoco}), and the reward changes based on the task (for example forward velocity or distance to a goal).

\subsection{Proximal Policy Optimization}

Although we expect any of the aforementioned reinforcement learning algorithms in Section~\ref{sec:intro} to learn effective skating maneuvers for locomotion (especially when we change the action space from continuous to discrete), for this paper we use the current state-of-the-art, Proximal Policy Optimization (PPO)~\cite{ppo}. In particular, PPO has achieved breakthrough results for continuous control robotics tasks by optimizing the following surrogate objective with clipped probability ratio: 
\begin{equation}
\label{ppo_obj}
L^{CLIP}(\theta) = \hat{\mathbb{E}}_t [\min(r_t(\theta)\hat{A}_t, \text{clip}(r_t(\theta),1-\epsilon,1+\epsilon)\hat{A}_t]
\end{equation}
where $\hat{A}_t$ is an estimator of the advantage function at time step $t$~\cite{Schulman15}, and $r_t(\theta)$ denotes the probability ratio

\begin{equation}
\label{ppo_prob_ratio}
r_t(\theta) = \frac 
			{{\pi}_{\theta}(a_t | s_t)} 
            {{\pi}_{\theta_{old}}(a_t | s_t)}
\end{equation}
where $\pi_\theta$ is a stochastic policy, and $\theta_{old}$ is the vector of policy parameters before the update.
This objective seeks to penalize too large of a policy update, which means penalizing deviations of $r_t(\theta)$ from 1.

\section{Modeling}
\label{sec:modeling}

As outlined in~\cite{skating2018icra}, planning effective, feasible skating motions for Robosimian involves two complementary problems. The motions of the skates must enable generation of required ground reaction forces without excessive slipping, to move the robot as desired, and solutions for inverse kinematics must be tractable, smooth, and within the dynamic velocity and acceleration limits of the joint actuators of the robot.

\subsection{Simple ``Representative'' System}

The first of the complementary problems is primarily focused on the skate locations, contact forces, and directions of motion. In fact, when designing the skating motions by hand, reasoning about the skate $(x,y,z)$ and yaw $(\phi)$ positions over time are the first things accounted for, and inverse kinematics are computed along this desired trajectory afterwards to ensure smoothness.

This gives rise to developing a more simple model to ``represent'' the full Robosimian model. As shown in the bottom left of Figure~\ref{fig_robosimianDev} and in Figure~\ref{img_action_space}, the model consists of the same torso/body, but without legs, which are replaced by floating bases above the skates.  The top rectangular link of the floating bases is moved with slide joints in MuJoCo~\cite{mujoco}, meaning motion is allowed along a single axis. Depending on the desired task, these can be actuated in the $x$, $y$, and/or $z$ directions in the floating base local frame.  The bottom rectangular link is actuated by a hinge joint about the $z$ axis, allowing yaw rotation only, which sets $(\phi)$ for the skate.  The total mass of one of Robosimian's limbs (excluding the skate) is distributed evenly between these two links. Although this simple system does not exactly model the true dynamics of the full Robosimian system, we hypothesize that it will be ``close enough'', making it faster to train a locomotion policy with deep reinforcement learning algorithms in Cartesian space. We hope to transfer this learned policy on the simple system onto the full system.

\subsection{Inverse Kinematics}

To use learned policies from the simple model, at run time (or for transfer learning and additional training on the full model), inverse kinematics (IK) must be used to map the desired skate motion back to the full model.

The IK to set the 6-DOF pose of the skate require choosing from among one of eight IK families, each analogous to a choice of ``elbow bending direction'' for each of three elbows on a limb~\cite{Byl14}.  Providing guarantees of smoothness requires precalculation of IK solutions across a region as well as compromises between ideal theoretical contact locations and achievable solutions for our particular robot.  In particular, for many desired skate configurations, achieving 
exact symmetry in end effector locations is either non-trivial or not achievable.

~\cite{Byl14} details algorithmic solutions for computing IK tables that satisfy the above conditions. However, as calculating such an IK table can be computationally expensive, 
there is a trade-off for training on the full model in joint space (without IK) vs. training in Cartesian space (making use of an IK table, or computing IK at each time step). 

Depending on the implementation, a function calculating IK can add significant overhead to training time in Cartesian space, as each time step requires 4 calls to this function (once for each limb). However, if the range of ($x$,$y$,$z$,$\phi$) of each skate in the simple model is limited to a subspace for which IK solutions of the full model both exist and are smooth, either through intuition or some pre-calculation, we can learn a policy on the simple model in Cartesian space, and then map the solution back at test time to the full system, computing IK at each time step for each limb. 
This eliminates the need to compute IK during training altogether.

\section{Training Environment}
\label{sec:details}

This section describes the environment set up and MDP details of our implementations to learn a locomotion policy for either the simple system or full Robosimian.

\subsection{Observation Space}
\label{sec:obs_space}

In order to use the policy trained on the simple model for the full model, the observation space (input to the network) must be the same, or similar. As there is no sensing available at the passive wheel in the real model, it is not fair to include any related observations to learn a policy in simulation. So neither the rotational position nor velocity of each skate wheel about its axis are included in the observation space.

At minimum, the observation space for the simple model consists of the following:
\begin{itemize}
\item ($x_b$, $y_b$, $z_b$), body global coordinates
\item ($w$, $x_w$, $y_w$, $z_w$), body orientation (from origin $x$-axis) in the form of a quaternion 
\item ($x_{s,i}$, $y_{s,i}$, $z_{s,i}$, $\phi_{s,i}$), skate local Cartesian positions and yaws, with respect to the body
\item ($dx_b /{dt}$, $dy_b /{dt}$, $dz_b /{dt}$), body translational velocities
\item ($d\theta_{x_b}/{dt}$, $d\theta_{y_b}/{dt}$, $d\theta_{z_b}/{dt}$), body rotational velocities
\item ($dx_{s,i}/{dt}$, $dy_{s,i}/{dt}$, $dz_{s,i}/{dt}$, $d\phi_{s,i}/{dt}$), skate local translational and rotational velocities w.r.t. body
\end{itemize}

The above observation space is (more than) enough to locomote the system in any given direction; 
for example to train for x-directed locomotion, the reward function can be a difference in potential between the current and previous body positions in the $x$ direction. 
For tasks that involve moving to a specific $(x_g,y_g)$ goal coordinate in the environment, the above observation space is augmented with the following: 
\begin{itemize}
\item ($x_g$,$y_g$,$z_g$), global goal coordinates
\item $-d$, negative of the absolute Euclidean distance between ($x_b$,$y_b$,$z_b$) and ($x_g$,$y_g$,$z_g$)
\item $\theta_{goal}$, local angle between current body heading ($w$, $x_w$, $y_w$, $z_w$) and ($x_g$,$y_g$,$z_g$)
\end{itemize}

The simple representative model and the full model using IK in Cartesian space thus have the same observation space in their respective simulations. The observations for which the real robot does not have direct sensing can readily be estimated with forward kinematics or Jacobians in real time on the real system.

When training in joint space for the full system, in addition to the body positions, orientation, and velocities, each joint's position and velocity (28 actuated motors) is now part of the observation space. Again, the rotational positions and velocities of the skate wheels are not included.

\begin{figure}[thpb]
      \centering
      \includegraphics[width=1.7in,height=1.265in]{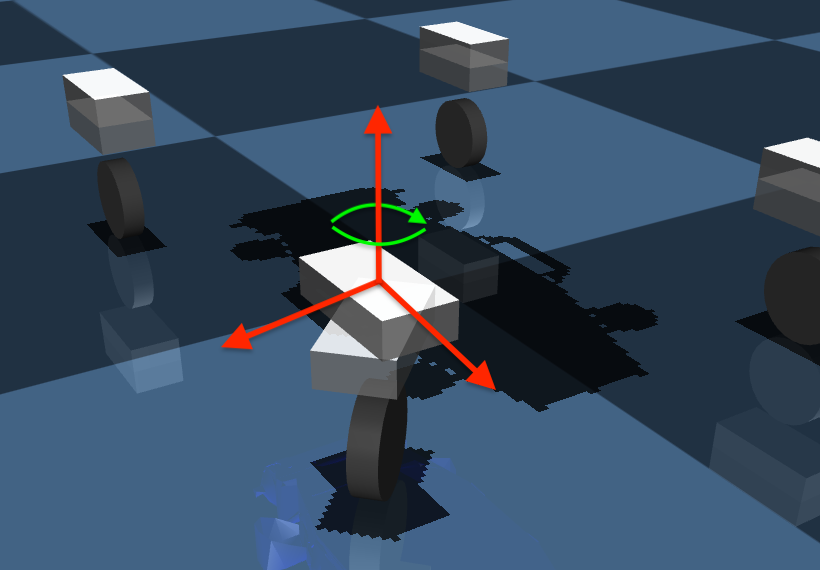}\includegraphics[width=1.7in]{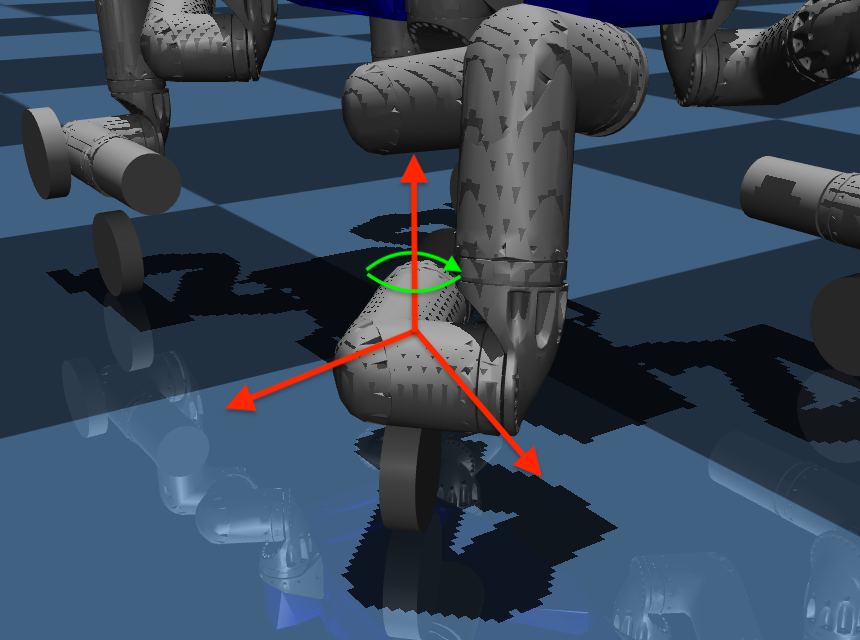}
      \caption{Equivalent states and action spaces for the simple Cartesian space model and full Robosimian model using IK. Actions are translation offsets along each skate's local $x$,$y$,$z$ axes, and rotational about the $z$ axis, shown here for $\phi=\pi/4$[rad].}
      \label{img_action_space}
\end{figure}
\subsection{Action Space}
\label{sec:action_space}

As discussed in Sections~\ref{sec:intro} and~\ref{sec:modeling}, it is more intuitive to model the actuators for Robosimian as position servos rather than torque motors. For the full system trained in joint space, the action space is chosen to be discrete, and the policy chooses an offset from $\{-\epsilon,0,+\epsilon\}$ from each motor's current desired position as its new reference.   

For the simple model and full model using IK, the action space is also discrete, but in Cartesian coordinates. This is shown in Figure~\ref{img_action_space}. At each time step, the policy chooses an offset from $\{-\epsilon,0,+\epsilon\}$ from each skate's ($x_{s,i}$, $y_{s,i}$, $z_{s,i}$, $\phi_{s,i}$) current desired positions. This $\epsilon$ is a design parameter, and can change based on the task and state. It is logical for $y_{s,i}$ and $\phi_{s,i}$ to change by different offsets $\epsilon_y$ and $\epsilon_{\phi}$, for example, due to the difference in units (meters vs. radians). A caveat is that the IK joint position differences of the full model between time steps must be bounded. We note that a small difference in the end effector location can have a large difference in the IK solution, even when using the same IK family, if near a singularity. 
We seek to minimize these events by keeping the simple model's workspace within smooth IK solution spaces for the full model. 

\subsection{Reward Functions}

We consider potential-based shaping functions of the form: 
\begin{equation} 
F(s,a,s') = \gamma \Phi(s') - \Phi(s)
\end{equation}
to guarantee consistency with the optimal policy, as proved by Ng et. al in~\cite{Ng_policy_invariance}. The real valued function $\Phi : S \rightarrow \mathbb{R}$ varies between tasks, with two such example tasks consisting of:
\begin{enumerate}
\item maximizing forward velocity in the $x$ direction:
\begin{equation}
\label{eq:fwd_vel}
\Phi(s) = \frac{x_b}{\Delta t}
\end{equation}

\item minimizing the distance to a target goal $(x_g,y_g,z_g)$:
\begin{equation}
\label{eq:min_dist}
\Phi(s) = - \sqrt{(x_b-x_g)^2 + (y_b-y_g)^2 + (z_b-z_g)^2}
\end{equation}
\end{enumerate}

This reward scheme gives dense rewards at each time step, towards ensuring the optimal policy is learned, and allows us to avoid complicated hand-crafted reward functions with many variables that ultimately output a single number anyway. Such schemes can result in slow training and sub-optimal behavior.

\subsection{Implementation Details}
We use a combination of OpenAI Gym~\cite{openaigym} to represent the MDP and MuJoCo~\cite{mujoco} as the physics engine for training and simulation purposes. We additionally use the OpenAI Baselines~\cite{baselines} implementation of PPO (PPO2) as a basis, making some key modifications for our system.
Our neural network architecture is the default Multi-Layer Perceptron (MLP), which consists of 2 fully connected hidden layers of 64 neurons each, with $\tanh$ activation. The policy and value networks both have this same network structure.

The default design parameters of these implementations are kept the same, while perhaps not producing the most optimal or time-efficient policies for our system, to show that intuitively reducing the action space has a large effect on training time and policy robustness.

\section{Results}
\label{sec:results}
We seek to compare the training times and robustness of learned policies for the following systems:
\begin{itemize}
\item \emph{SS}: Simple ``representative'' system
\item \emph{FS in JS}: Full system trained in joint space
\item \emph{FS in CS}: Full system trained in Cartesian space with IK (to set joints)
\end{itemize}

We also seek to evaluate how well the learned policy of the simple system transfers to the full system with inverse kinematics. \emph{SS} and \emph{FS in CS} always have the same observation and action spaces, for fair comparisons. We consider tasks of skating at maximum velocity in the $+x$ direction, and of locomoting to a particular goal location $(x_g,y_g)$. All policies along with additional comparisons to other control methods are shown in the supplementary video for this paper.

\subsection{Skate Straight}
First we consider the task of maximizing forward velocity in the $+x$ direction. The observation and action spaces are as detailed in Sections~\ref{sec:obs_space} and~\ref{sec:action_space}, with $\epsilon=0.01$. The action space for \emph{SS} and \emph{FS in CS} is limited to a $\pm 0.1$[m] offset of $y_{s,i}$ and $\pm 0.3$[rad] offset of $\phi_{s,i}$ for each skate from it's given starting position, with $x_{s,i}$ and $z_{s,i}$ fixed, to match the constraints used in designing skating motions for forward locomotion in~\cite{skating2018icra}. The action space for \emph{FS in JS} is bounded only by the joint limits. Training on \emph{FS in JS} also enforces early termination of an episode if any self-collisions are detected, or if contact occurs with the ground from any part of the robot other than the skates.
The reward at each time step is a potential-based shaping function where $\Phi$ is as in Equation~\ref{eq:fwd_vel}, resulting in:
\begin{equation}
F(s,a,s') = \frac{x_b' - x_b}{\Delta t},
\end{equation}
rewarding body velocity in the $+x$ direction. 

\subsubsection{Training Sample Complexity Comparison}
\begin{figure}[thpb]
      \centering
       \includegraphics[width=3.0in]{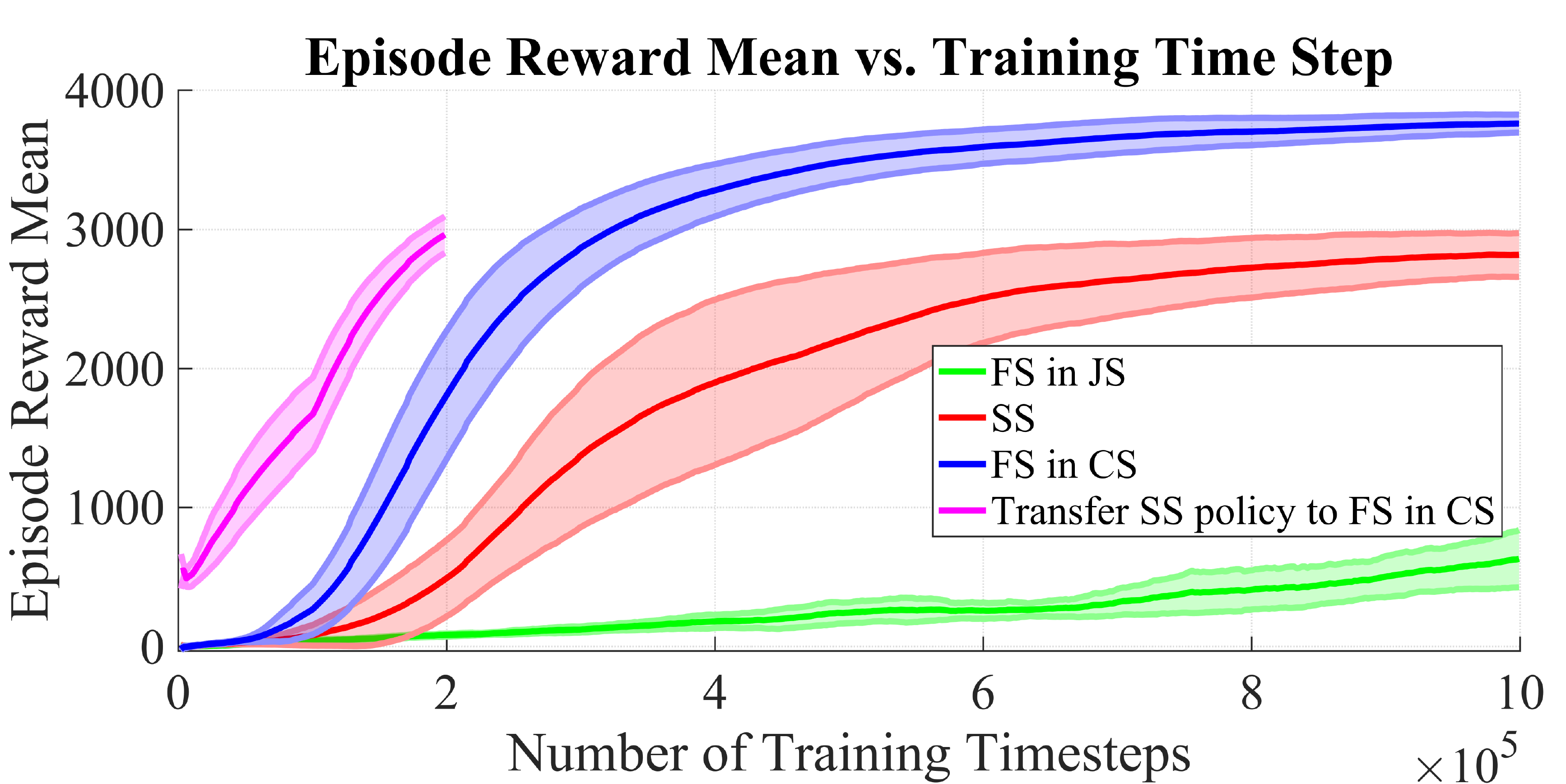} \\ 
      \caption{Average episode reward over training 1 million time steps for the full system in joint space (\emph{FS in JS}), simple system (\emph{SS}), and full system in limited Cartesian space (\emph{FS in CS}) for a task rewarding forward velocity in the $+x$ direction. 
      The pink line shows the results of transferring the learned policy on the simple system (\emph{SS}) to the full system in Cartesian space (\emph{FS in CS}) and further training for 2e5 timesteps. As the dynamics do not match perfectly, using the \emph{SS} can be a means to accelerate training, especially if fast IK computation during training is not available.
      }
      \label{fig_rew_mean_training_timestep}
\end{figure}

\begin{figure}[thpb]
      \centering
       \includegraphics[width=2.95in]{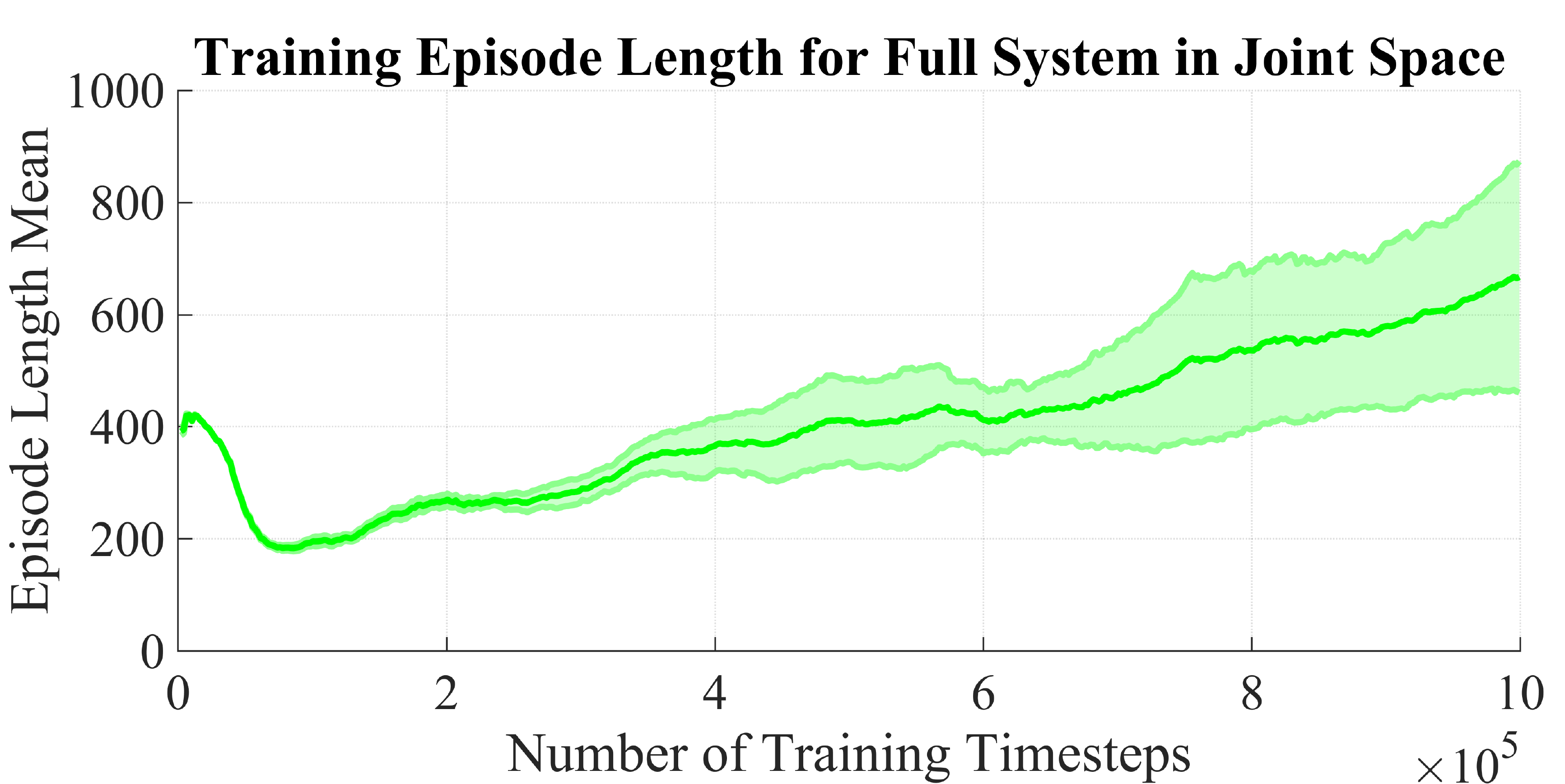} \\ 
      \caption{Episode length mean vs. number of training time steps, where each episode normally lasts 1000 training time steps unless ended prematurely. 
      Early termination is the result of any internal self-collisions, or any part of the robot body touching the ground other than the skates, which is still occurring even after training for over 1e6 timesteps.
      }
      \label{fig_eplen}
\end{figure}

Figure~\ref{fig_rew_mean_training_timestep} shows the episode reward mean vs. number of training time steps, with mean and standard deviation of three training runs for each system. The episode reward is the sum of the individual rewards at every time step, so it is the sum of all instantaneous velocities, a maximum of 1000 values (without early termination). We see that training in Cartesian space gives much higher total returns, with far fewer time steps. This is likely due to both the smaller observation and action spaces for \emph{SS} and \emph{FS in CS}, as well as to the fact that the agent for the full system is being forced to learn forward and inverse kinematics in conjunction with trying to maximize returns.
A training episode terminates early if there are any self-collisions in the current robot configuration, or if non-skate limbs come in contact with the ground, where the average duration of a training episode before one of these occurrences is shown in Figure~\ref{fig_eplen}. 

In addition to learning a policy that produces returns that are not nearly as high as those for \emph{SS} and \emph{FS in CS}, the resulting motions for training on \emph{FS in JS} also do not \emph{look} optimal, where typical behavior appears to be sinking quite low to the ground (always on the brink of a collision) and waving one of it's rear limbs around in the air while only the other three limbs actually skate (see video). Even after training for 10 million time steps, which takes about 12 hours on a single laptop, the policy has still not completely learned kinematics, frequently terminating episodes early, and it is still not achieving the same returns as the policy trained in Cartesian space.

\subsubsection{Training Wall Clock Time Comparison}
\begin{figure}[thpb]
      \centering
      \includegraphics[width=2.95in]{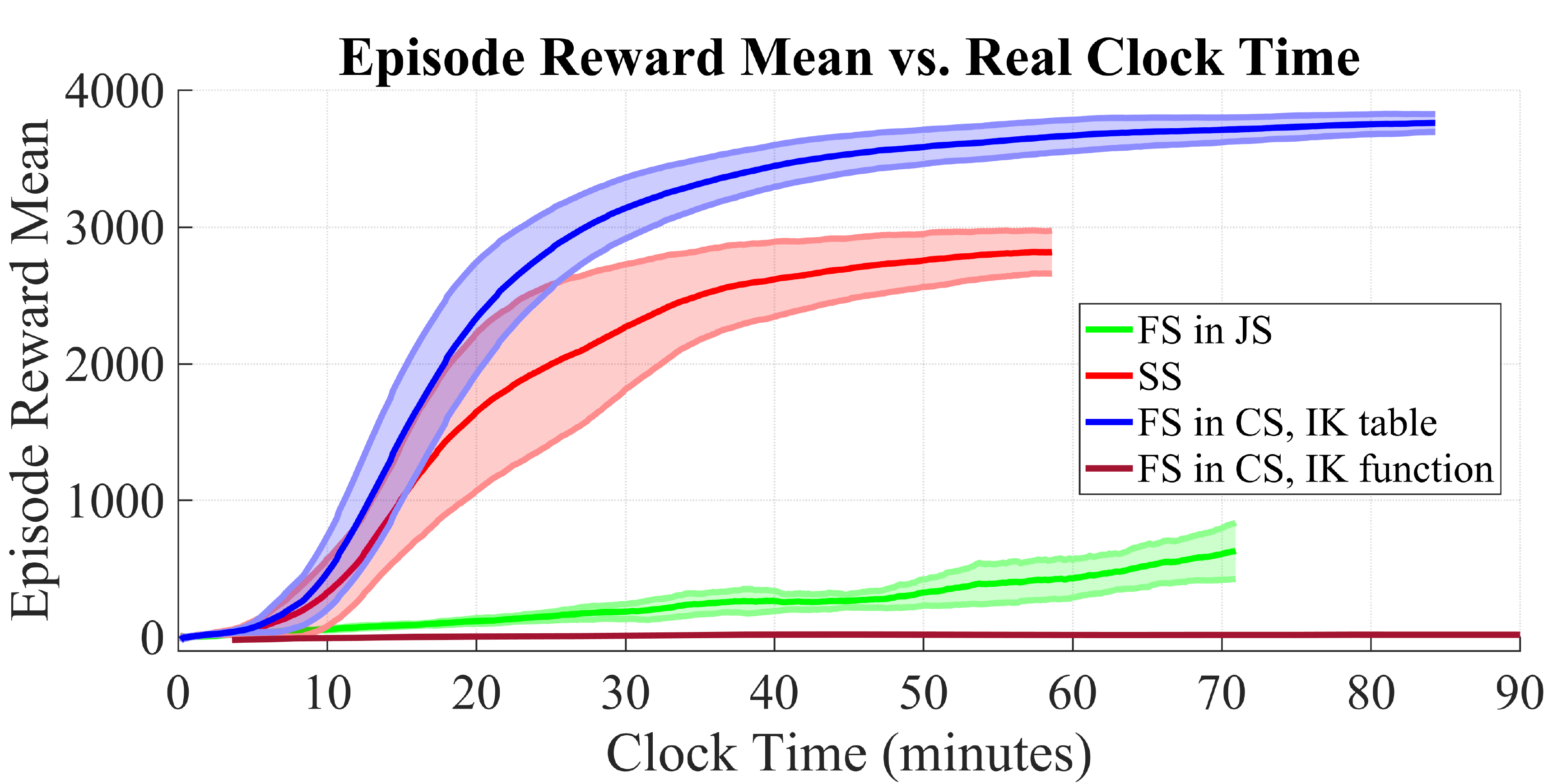} \\
       \includegraphics[width=2.95in]{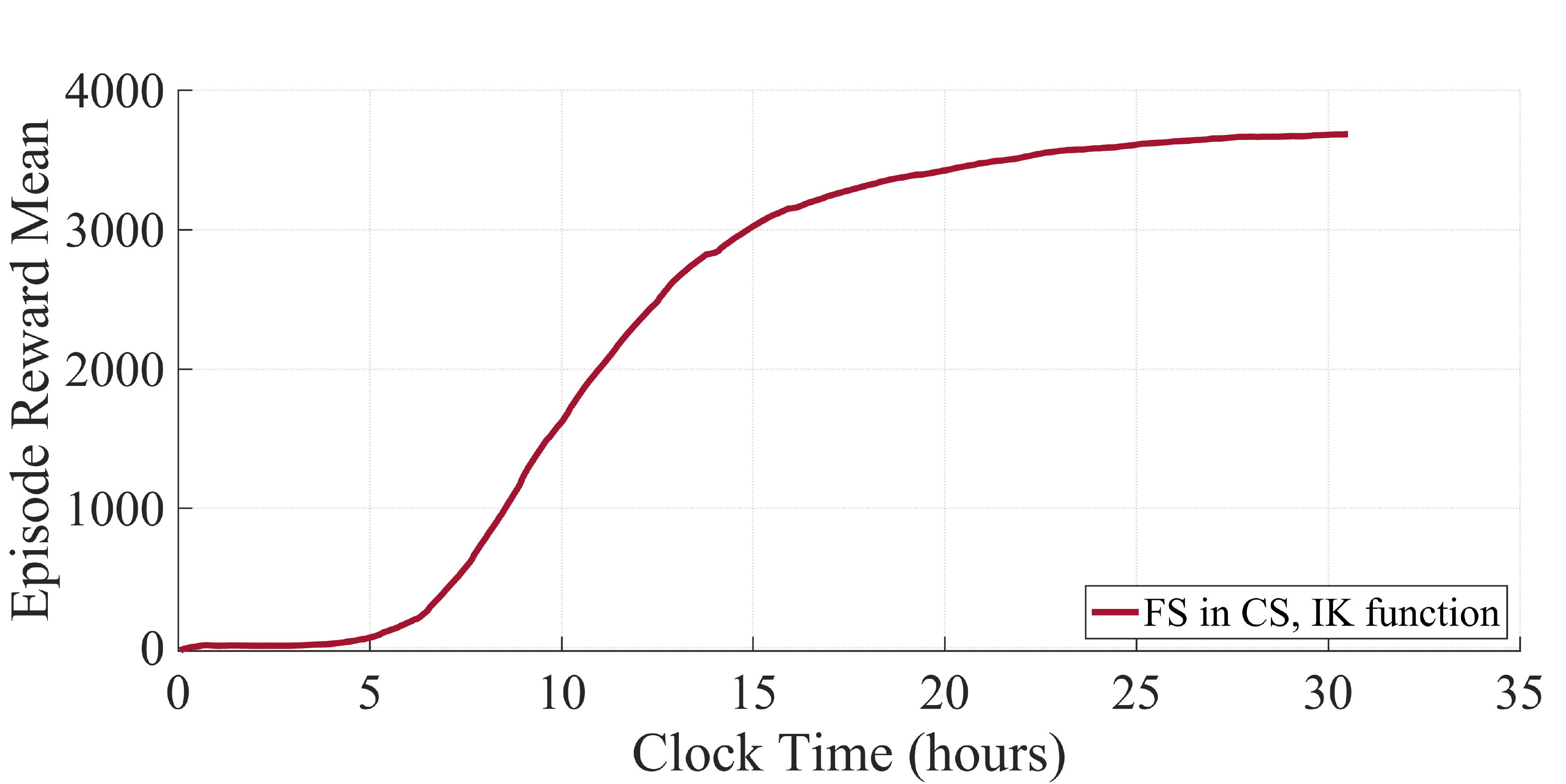} \\ 
      \caption{Episode reward mean vs. real clock time for training over 1 million time steps. The \emph{SS} is fastest to train due to the fewest number of weights to learn in the network, limited search space (along with \emph{FS in CS}), and the simplest dynamics to simulate in MuJoCo. If an IK table is available for use, the overheard of looking up values does increase training time, but this overhead is not significant compared to using the IK function during training (4000 times per training episode!).
       }
      \label{fig_rew_mean_real_time}
\end{figure}

Figure~\ref{fig_rew_mean_real_time} shows the episode reward mean vs. wall clock time. The overhead in training time for the $FS$ in $JS$ compared to the $SS$ can be attributed to more weights in the neural networks having to be learned due to the larger observation and action spaces, more frequent environment resets from early terminations,
and possibly also from simulating more complex dynamics. If an IK table (implemented as a hash table, where the keys are $(x,y,z,\phi)$ position tuples and values are the corresponding joint positions) is available during training time, this adds a small overhead from 4 constant time look ups per training time step. Without such a table, calculating IK 4 times at each time step introduces a very significant overhead. 
Without a fast method for computing IK, this may suggest transferring a policy learned on the \emph{SS} onto the \emph{FS in CS}, calculating IK only at run time.

An IK table can also be dynamically generated as training progresses, to avoid frequently re-calculating the same position tuples as the policy learns subspaces of the workspace that maximize rewards. We also note that the training time from slow IK calculations is still worth it, as the alternative of training in joint space produces less robust policies that attain far fewer rewards.

\subsubsection{Policy Transfer from \emph{SS} to \emph{FS in CS}}

To test how well the policy learned on the simple model transfers to the full system, we initialize the weights of the policy and value networks for the \emph{FS in CS} with those learned from the \emph{SS}. Since the observation and action spaces are the same for both models, we do not run into any implementation complications.
The pink line in Fig. 
~\ref{fig_rew_mean_training_timestep} shows the episode reward mean for training $2\text{e}5$ time steps after initializing the \emph{FS in CS} network weights with those learned with the \emph{SS}.

The initial returns start much higher than 0, but not as high as the returns from running the \emph{SS} policy on its own system. This is expected since the dynamics do not match perfectly, but it is encouraging that locomotion can be transferred in this manner. This suggests the possibility of training on the simple system for other tasks, and then executing the learned policy on the full system, to accelerate training.

A noteworthy observation from our experiments is that the learned policy for the \emph{SS} could not traverse randomly varying, sinusoidally-smooth terrain, as the amplitude increased, as defined in Section~\ref{sec:skate_to_goal}. 
This could be due to the dynamics of the system being incapable of generating the forces to traverse this new terrain, or that this new observation space had not been explored during training, and thus the policy had not learned which actions to take in these slightly varying states. However, the full robot \emph{was} able to make forward progress after transferring that same \emph{SS} policy onto the \emph{FS in CS}, without further training of that policy. This implies 
that the policy was robust enough, 
but the \emph{SS} version of the system itself may not have the means to produce the forces necessary to get over the small hills. This further encourages the use of training a policy on the \emph{SS} and then transferring it to the full system for other tasks.

\subsection{Skate To Goal Under Uncertainty}
\label{sec:skate_to_goal}

The next task we consider is locomoting the system from the origin (0,0) to a goal location $(x_g,y_g)$, in particular (5,0)[m], over randomly varying, sinusoidally-smooth terrain with varying friction coefficients. For this task, the observation spaces from the maximum velocity skating task are augmented with the global goal coordinates, negative distance to the goal from the robot's current location, and the angle between the robot's heading and the goal, as discussed in the latter part of Section~\ref{sec:obs_space}. The action spaces are left unchanged, as the limited spaces for \emph{SS} and \emph{FS in CS} should be enough to locomote the system to the goal (it is always possible to increase ranges, or allow $x$ and $z$ skate position changes if the terrain is \emph{too} rough).
The reward at each time step is again a potential-based function, with $\Phi$ as in Equation~\ref{eq:min_dist}, where we assume $z_g = z_b$, resulting in:
\begin{equation}
\begin{split}
F(s,a,s') = - \sqrt{(x_b'-x_g)^2 + (y_b'-y_g)^2} \\ - 
(- \sqrt{(x_b-x_g)^2 + (y_b-y_g)^2})
\end{split}
\end{equation}
rewarding a decrease in the distance to the goal, and penalizing moving away from the goal.

We train the policy to reach the goal (5,0) for 2 million time steps over smooth, sinusoidal terrain with amplitude $A=0.1$ [m], period $2\pi$ [m] in both $x$ and $y$ directions, and coefficient of friction $\mu \in [0.5,1]$. The terrain is randomly generated and moved in the $xy$ plane by ($\delta x, \delta y) \in [-1,1]$ [m] at the start of each training episode (environment reset), on top of random perturbations of the joints, positions, and velocities of the initial states. An episode is considered completed when ($x_b$,$y_b$) is within 0.2 meters of ($x_g$,$y_g$), or when the episode times out after 1000 time steps.

We then test the policy over the same family of smooth sinusoidally varying terrain now also varying amplitude $A \in [0,0.2]$ [m] (the period $2\pi$ [m] and coefficient of friction range $\mu \in [0.5,1]$ are unchanged), and compare results with runs of our previously designed open-loop trajectory to locomote the system 5 meters forwards from~\cite{skating2018icra}.

\begin{figure}[thpb]
      \centering
       \includegraphics[width=3.0in]{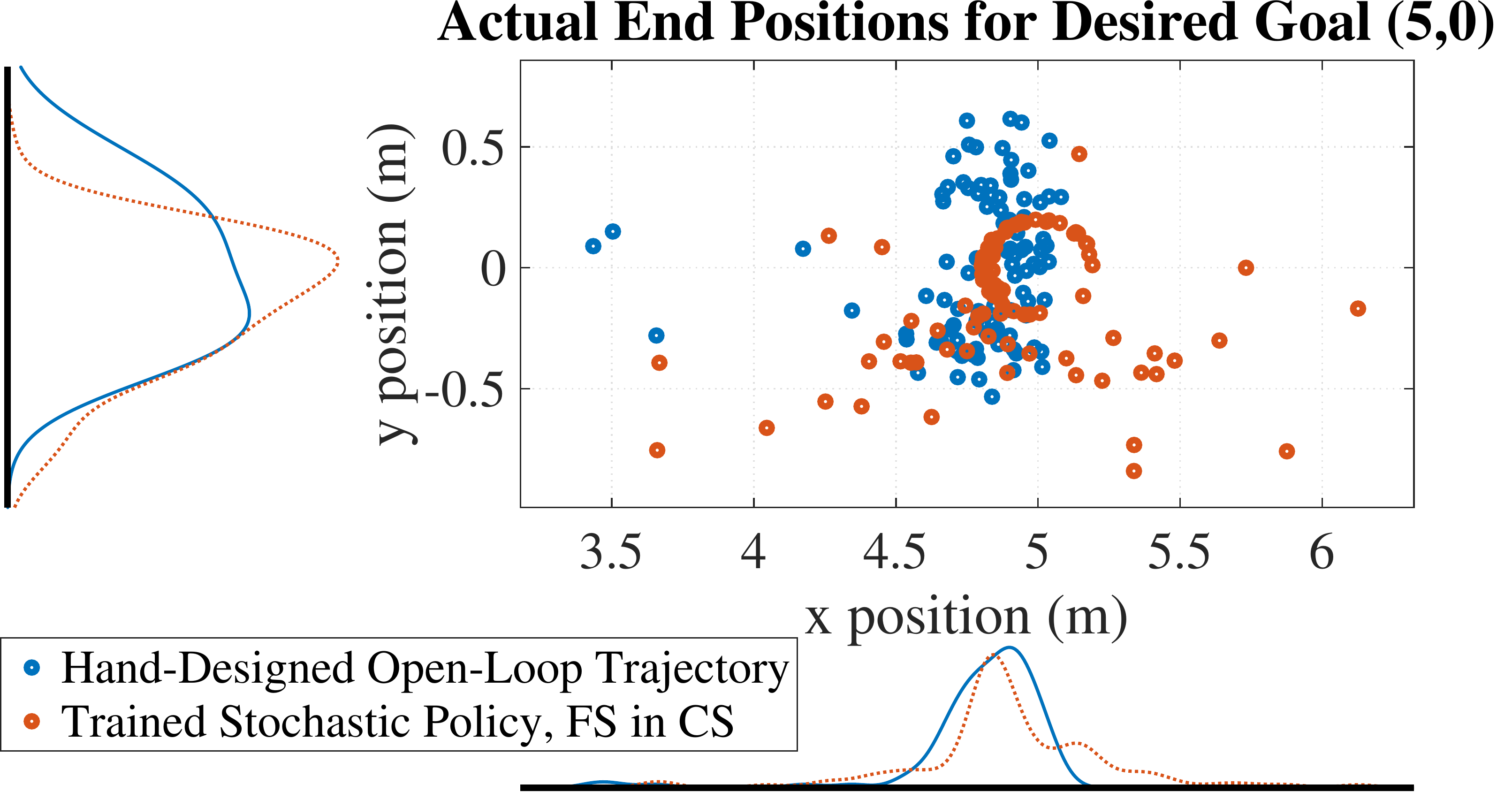}\\
      \caption{ Starting from the origin and with a goal of reaching (5,0) in the environment, resulting end positions for the robot body center of mass from 100 runs each of the learned policy on the \emph{FS in CS} as well as of a hand-designed trajectory. The terrain consists of randomly varying smooth, sinusoidal curves with amplitude $A\in[0,0.2]$ [m], period $2\pi$ [m], and coefficient of friction $\mu \in [0.5,1]$.
      }
      \label{fig_end_pos}
\end{figure}

Figure~\ref{fig_end_pos} shows the end $(x_b,y_b)$ positions and distributions for 100 runs each for the trained stochastic policy on the \emph{FS in CS} as well as for the hand-designed trajectory.
Reaching the goal is defined as successful if ($x_b$,$y_b$) ends within 0.2 meters of ($x_g$,$y_g$), 
for both methods. The large cluster of policy points in a radius of 0.2 meters around (5,0) is due to the episode completion condition.  
Of the 100 trials, there are 57 successes using the learned policy, vs. only 23 successes for the hand-designed open loop trajectory. 
From the resulting end positions and observing the policy in action, we see that if the robot has too much lateral error, due to the small action space ranges, it cannot move laterally towards the goal and gets ``stuck''. An even more robust policy might thus be learned by increasing the action space ranges, and incorporating more knowledge into the observation space such as sensor readings of terrain height, for future work.

\subsection{Policy and Hand-Designed Trajectory Comparison}

\begin{figure}[thpb]
      \centering
       \includegraphics[width=2.9in]{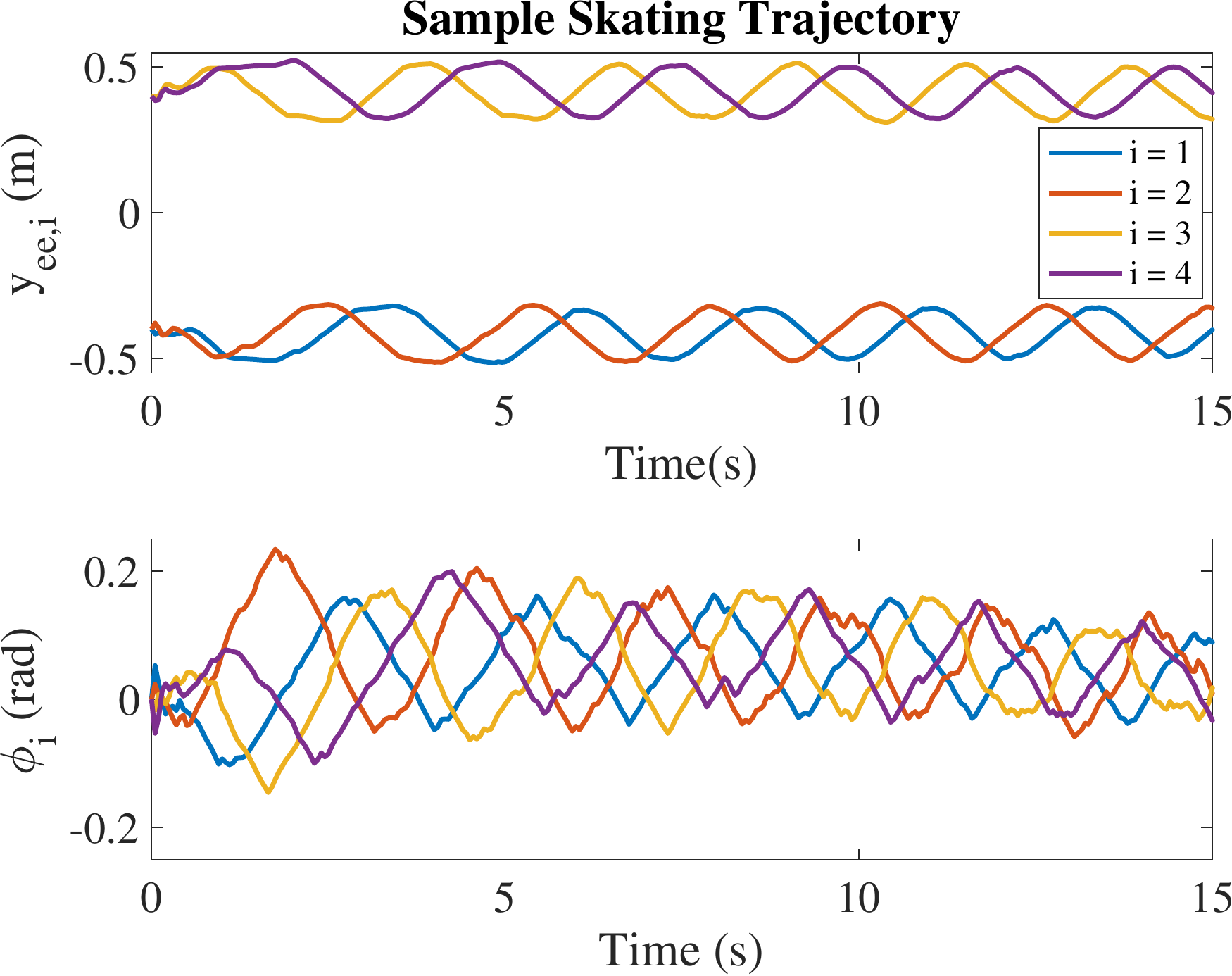} \\
      \caption{The first 15 (s) of a sample skating trajectory from a policy learned for the full system in Cartesian space, after training for 1 million time steps while rewarding forward velocity in the $+x$ direction.  When viewing the robot from above, skate $i=1$ is on the front right limb, $i=2$ is the rear right, $i=3$ is on rear left, and $i=4$ is the front left.
      }
      \label{fig_rl_skate_results}
\end{figure}

The first 15 (s) of a sample skating trajectory from a policy trained to achieve maximum velocity in the $x$ direction for the \emph{FS in CS} are shown in Figure~\ref{fig_rl_skate_results}. 
The resulting motions are quite similar to our hand-designed trajectories in~\cite{skating2018icra}, with symmetric motions in both $\phi$ and $y$ position between the left and right limbs, as well as an approximate phase difference of $\pi/2$ between the front and rear limbs.
      There are two key differences: (1) although $\phi$ is allowed to vary in $[-0.3,0.3]$ radians, the policy learns trajectories that never reach those limits, and (2) as the robot builds speed, the $\phi$ offsets chosen decay to a smaller range and oscillate more quickly, along with $y$, once slip is avoided at the start of the motion.
The non-smooth $\phi$ motions could be due to the nature of 
executing a stochastic policy, or that the policy has learned to make small corrections in its heading with $\phi$, rather than with $y$, to maximize forward velocity reward.

\section{Conclusion}
\label{sec:conclusion}
This paper details the use of targeting state-of-the-art reinforcement learning algorithms on simple, representative systems with standard control techniques, and intuitively limiting the action space to reduce sample complexity, increase robustness, and accelerate training time. 

We showed that using inverse kinematics and training in Cartesian space significantly speeds up training time.
Even without a fast way to compute IK, the resulting policies are more stable and robust than training in joint space, where even after millions of time steps, the agent is still learning kinematics, producing non-intuitive motions, and terminating training episodes early. Such considerations are especially important for a high DOF system such as Robosimian.

We also provide a workaround in the case of no fast IK being available, by training a policy on a simple representative system, and transfer learning onto the full system, calculating IK only at run time, or for fewer time steps if further training of the policy is needed due to mismatching dynamics.

Although it may seem intuitive to use inverse kinematics to learn on a lower dimensional system, to the best of the authors' knowledge, this is the first such work in this area. Here this is applied to a skating system, but the ideas can be readily implemented on any system with a base and end effector(s). 
Example applications could include moving a manipulator to grasp an object in 3D space, or a quadruped placing down a foot during a particular gait.

\bibliographystyle{IEEEtran}
\bibliography{root_arxiv.bbl}

\end{document}